\documentclass[letterpaper, 10 pt, conference]{ieeeconf}  
\IEEEoverridecommandlockouts                              

\usepackage{graphics} 
\usepackage{graphicx}
\usepackage{subfigure}
\usepackage{multirow}
\usepackage{mathptmx} 
\usepackage{amsmath} 
\usepackage{amssymb}  
\usepackage{textcomp}
\usepackage{soul} 
\graphicspath{{figures/}}
\usepackage{cite}

\author{Egor Lakomkin$^{*1}$, Mohammad Ali Zamani$^{*1}$, Cornelius Weber$^{1}$, Sven Magg$^{1}$ and Stefan Wermter$^{1}$
\thanks{*indicates equal contribution}
\thanks{$^{1}$University of Hamburg, Department of Informatics, Knowledge Technology Institute. 
Vogt-Koelln-Strasse 30, 22527 Hamburg, Germany
        {\tt\small $\{$lakomkin, zamani, weber, magg, wermter $\}$ @informatik.uni-hamburg.de }}%
}
\title{\LARGE \bf
EmoRL: Continuous Acoustic Emotion Classification using Deep Reinforcement Learning
}

\begin{document}

\maketitle
\thispagestyle{empty}
\pagestyle{empty}

\begin{abstract}

Acoustically expressed emotions can make communication with a robot more efficient. Detecting emotions like anger could provide a clue for the robot indicating unsafe/undesired situations. Recently, several deep neural network-based models have been proposed which establish new state-of-the-art results in  affective state evaluation. These models typically start processing at the end of each utterance, which not only requires a mechanism to detect the end of an utterance but also makes it difficult to use them in a real-time communication scenario, e.g. human-robot interaction. We propose the EmoRL model that triggers an emotion classification as soon as it gains enough confidence while listening to a person speaking. As a result, we minimize the need for segmenting the audio signal for classification and achieve lower latency as the audio signal is processed incrementally. The method is competitive with the accuracy of a strong baseline model, while allowing much earlier prediction.

\end{abstract}

\section{INTRODUCTION}
 \par Emotions are essential for natural communication between humans and have recently received growing interest in the research community. Dialog agents in human-robot interaction could be improved significantly if they were given the ability to evaluate an emotional state of a person and its dynamics. For instance, if a robot could detect that a person is speaking in an angry way which could be a sign that the robot should adjust its behavior. 
 
 \par  Deep Neural Networks (DNN) have been successfully applied in speech and natural language processing tasks such as language modeling \cite{jozefowicz_exploring_2016}, sentiment analysis \cite{radford_learning_2017}, speech recognition \cite{hannun_deep_2014} and neural machine translation \cite{wu_googles_2016}. DNNs have also been adapted for emotion recognition problems falling into three main categories: frame-based processing \cite{fayek_evaluating_2017} (usually with a majority voting for the final classification)  and sequential processing \cite{huang_attention_2016} (taking into account the temporal dependencies of the acoustic signal) or a combination of both \cite{lee_high-level_2015}. 
 
 The objective of this previous work is to achieve the highest possible classification accuracy 
 given the entire utterance. Usually an extra mechanism is required to detect the end of the utterance to make a prediction, but in reality humans can evaluate an emotional state of a person already before a phrase or sentence is finished. Existing models rely on acoustic segmentation methods to detect speech boundaries which are then classified by the model. Also, Bi-directional Recurrent Neural Networks (Bi-RNN) have been shown to significantly outperform forward-only architectures in  sequence classification \cite{longpre2016way}.
 As a disadvantage, Bi-RNNs can only process the utterance when it is finished which is not desired in situations like real-time processing or safety-related issues. For instance, analyzing just one second of the utterance instead of the entire utterance can determine whether a person is in a highly negative state, which could lead to a crucial time margin for safety reasons.
 
 \begin{figure}[t]
  \includegraphics[width=\linewidth]{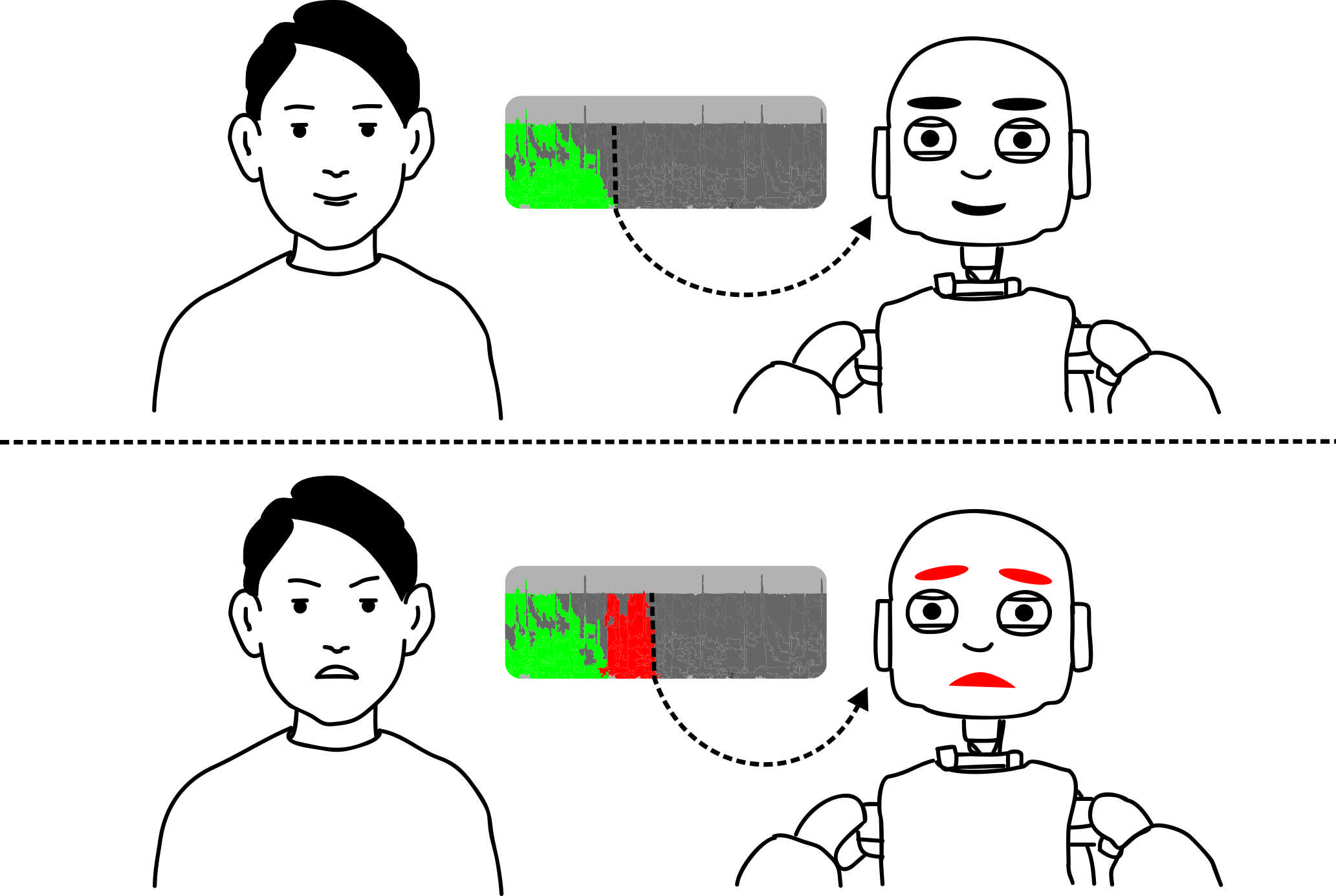}
  \caption{High-level system overview. The robot analyzes continuously arriving acoustic input and only when it has enough information to evaluate the affective state of the speaker it will output if the person is in an angry state or not. The model evaluates audio input every 300ms and also takes into account information from the past. An agent  is trained using reinforcement learning to make the dynamic decision: wait for more audio data or trigger prediction now. Please refer to the supplementary video.}
  \label{fig:system_main}
  \vspace{-30pt}
\end{figure}

\par In this paper, we propose a system that learns to perform emotion classification, which optimizes two factors: accuracy and latency of the classification. We cast this problem into a reinforcement learning problem by training an emotion detection agent to perform two actions: \textit{wait} and \textit{terminate}. By selecting the \textit{terminate} action, the agent stops processing incoming speech frames and performs classification based on the information it observed. A trade-off between accuracy and latency is achieved by punishing wrong classifications as well as too late predictions through the reward function.

\par Our main contribution is a neural architecture that learns to predict an emotion state of a speaker with minimum possible latency without significantly sacrificing  the accuracy of the system. Our model is especially useful for robot applications, for example, to detect an unsafe situation earlier given a human utterance. We evaluate the proposed model on the iCub robot platform and compare it with multiple baseline models. 

\section{RELATED WORK}
\par The majority of previous work in acoustic emotion recognition focuses mainly on utterance-level classification. The structure of the available annotated data is one of the reasons: labels are provided for the whole acoustic signal. For instance, there could be a sample of 5-6 seconds length where several emotions are mixed together, or there might be even several speakers expressing different emotions, but there is no information about boundaries or time frames. As a result, it is common practice to assume that the annotation label corresponds to the whole content of the utterance. In previous research, two main directions can be observed: to model emotion for the whole utterance directly or to model emotion of a short acoustic chunk and combine individual predictions to infer a label for the whole sequence.

\par \textbf{Utterance-level classification}: Recurrent architectures model long-term temporal dependencies and are a popular choice to model the sequence-level emotion classification.  Huang et al. \cite{huang_attention_2016} proposed an attention-based recurrent neural network, which implicitly learns which speech frames are important for the predictions as there could be a significant amount of non-relevant information, such as silence. Learning task-specific representations directly from the raw data using neural networks has recently gained popularity, mainly inspired by the success of convolutional neural networks in computer vision problems \cite{krizhevsky_imagenet_2012,simonyan_very_2014}. Ghosh et al. \cite{ghosh_representation_2016} demonstrated a successful example of using autoencoders for feature learning from  power spectrograms and RNN pre-training. Trigeorgis et al.
proposed an architecture which combines convolution and recurrent neural networks to learn features from a raw waveform for emotion classification \cite{trigeorgis_adieu_2016} instead of using well-known hand-crafted representations like Mel-Frequency Cepstral Coefficients (MFCCs).

\par \textbf{Frame-level classification}: As an alternative to utterance-level modeling, finer-grained emotion classification could be feasible with speech frames which are not labeled the same.
Fayek et al. \cite{fayek_evaluating_2017} achieved state-of-the-art performance with convolutional neural networks modeling a probability distribution over emotion classes at the speech frame level and selecting the emotion with the highest average posterior probability over all frames as a final decision. Lee et al. \cite{lee_high-level_2015} obtained an aggregated vector by the frame-level predictions with several statistic functions and fed the vector to an Extreme Learning Machines classifier. 

\par  \textbf{Adaptive sequence processing}: A recently emerging area of natural language processing is based on a combination of reinforcement learning with traditional supervised settings. 
Yu et al. \cite{P17-1172} proposed a variant of the Long-Short Term Memory model \cite{hochreiter_long_1997} which is able to skip irrelevant information by deciding how many forthcoming words can be omitted. The model is trained with the REINFORCE algorithm \cite{williams1992simple} and it showed that it needed to process 1.7 times less words to achieve a similar accuracy level as an LSTM processing the whole sentence in a sentiment analysis task. Shen et al. \cite{shen_reasonet:_2017} achieved state-of-the-art performance in machine comprehension by introducing a termination gate as an additional LSTM gate, which is responsible for an adaptive stop. A combination of the optimization of cross-entropy loss and expected reward was proposed by Ranzato et al. \cite{ranzato_sequence_2016} which allowed to train models with a large action space, for example, in the text generation domain \cite{gu_learning_2017}.


\par Our proposed EmoRL model, is inspired by recent advances of adaptive sequence processing architectures \cite{P17-1172}, \cite{ranzato_sequence_2016} and\cite{shen_reasonet:_2017}, and learns how to terminate and classify the emotion as early as possible. To our best knowledge, this is the first example of such a model in the acoustic signal processing domain.

\section{METHODOLOGY}


Given the sequence of utterances, EmoRL, our proposed model, can determine the earliest reasonable time to classify an emotion. EmoRL receives the acoustic features as a raw state of the environment. As can be seen in Fig.\ref{fig:system_architectures}, we divide our proposed model into three parts, the GRU for the state representation, the emotion classification and the action selection module. Since each frame length is 25ms, which is too short to detect the underlying emotion, multiple frames are necessary to achieve a more descriptive state ($\theta_s$). A temporal abstraction of given features, which is already provided in the Gated Recurrent Unit (explained in section III.B), is a more efficient state representation. This state is shared with both the emotion classification ($\theta_c$) and the action selection module ($\theta_a$) which determines when to terminate listening to the utterance.

\subsection{Feature extraction}

We extract 15 MFCC coefficients and their first and second derivatives extracted from windows of 25ms width and 10ms stride using the OpenSMILE toolkit \cite{eyben_recent_2013}. In addition to MFCC coefficients we extract fundamental frequency values (pitch), voice probability and loudness smoothed with a moving average window with a size of 15. The reason for our choice of features is that such feature set showed state-of-the-art results in acoustic emotion classification by Huang et al. \cite{huang_attention_2016}. 
We normalize each feature based on mean and standard deviation statistics calculated over the training dataset. Each feature is subtracted with the mean and then divided by the standard deviation.

\subsection{Emotion Classification Model}
\begin{figure}[t]
\vspace{5pt}
  \includegraphics[height=12cm, width=8.5cm]{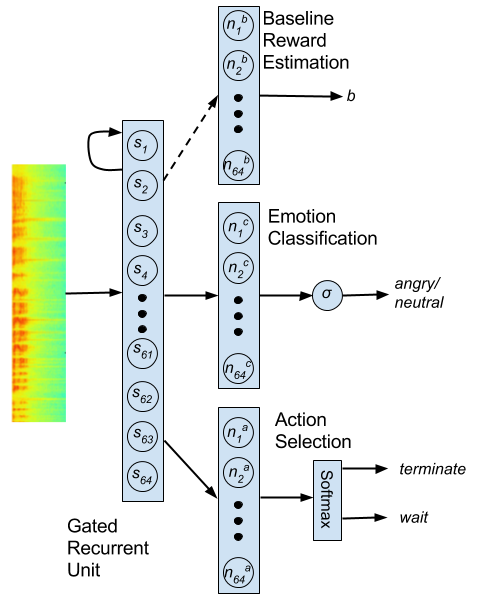}
  \caption{The EmoRL model consists of 4 components: \textit{Gated Recurrent Unit (GRU)}, \textit{Emotion Classification (EC)}, \textit{Action Selection (AS)} and \textit{Baseline Reward Estimator (BRE)}. The GRU encodes the acoustic information as a fixed-length vector allowing to model long-term dependencies of a speech signal which is used as a state representation in our system. \textit{EC} is a single layer module which uses the state representation to evaluate the probability of the human speaker being in an angry state. \textit{AS} and \textit{BRE} are also single layer modules which determine the probability distribution over possible actions and the estimation of the baseline reward.}
  \label{fig:system_architectures}
  \vspace{-20pt}
\end{figure}

For the emotion classification model (see Fig.\ref{fig:system_architectures}), we use a single layer Recurrent Neural Network with Gated Recurrent Units (GRU) proposed by Bahdanau et al. \cite{bahdanau_neural_2015}. The GRU network updates its internal memory state at each time frame. We average all hidden memory states to obtain a compact fixed-length vector representation of the utterance which we feed to the classification layer. We select this part of the model ($[\theta_s;\theta_c]$) (indicated as GRU\_Baseline in this paper) to compare it with our EmoRL, due to its simplicity, and since it was demonstrated by Huang et al. \cite{huang_attention_2016} that such architecture produces state-of-the-art results.

\setlength{\belowdisplayskip}{0pt}
\setlength{\belowdisplayshortskip}{0pt}
\setlength{\abovedisplayskip}{0pt} 
\setlength{\abovedisplayshortskip}{-3pt}

In each time frame, the extracted features $x_t \in \mathbb{R}^{33}$ are passed to the GRU layer to obtain the hidden memory representation.  
\begin{equation}
\label{eq.gruhidd}
z_t = \sigma (x_t U^z + s_{t-1} \cdot W^z)
\end{equation}

\begin{equation}
\label{eq.gruhidd2}
r_t = \sigma (x_t U^r + s_{t-1} \cdot W^r) 
\end{equation}

\begin{equation}
\label{eq.gruhidd3}
h_{t} = \tanh(x_t U^h + (s_{t-1} \odot r_{t} ) W^h)
\end{equation}

\begin{equation}
\label{eq.gruhidd4}
s_{t} = z_t \odot s_{t-1} + (1 - z_{t} )\odot h_{t-1}
\end{equation}
where $z$ and $r$ are update and reset gates, $s_t$ is the memory representation at time frame $t$, $U$ and $W$ are parameters matrices and $\sigma (\cdot)$ is the logistic sigmoid function. Then, the emotion is determined by the next layer:
\setlength{\belowdisplayskip}{3pt}
\setlength{\belowdisplayshortskip}{0pt}
\setlength{\abovedisplayskip}{3pt} 
\setlength{\abovedisplayshortskip}{0pt}
\begin{equation}
\label{eq.dec1}
d_{t} = \sigma(w^d\cdot S_t + b^d)
\end{equation}

\begin{equation}
\label{eq.dec2}
emotion = \left\{\begin{matrix}
angry & d_t>0.5 \\ 
 neutral & d_t \leq  0.5 
\end{matrix}\right.
\end{equation}
We use a binary cross-entropy loss function to train the emotion classification model:

\setlength{\belowdisplayshortskip}{3pt}
\setlength{\abovedisplayshortskip}{0pt}

\begin{equation}
\label{eq.dec1}
J_{c} (\theta_s, \theta_c) = - \hat{d} \ \log d_T + (1-\hat{d}) \ \log(1-d_T)
\end{equation}
where $\hat{d}$ is the ground truth for the emotion (0 for neutral and 1 for angry) and  $d_T$ the predicted emotion at the final time frame (or the terminal time frame). $\theta_s$ and $\theta_c$ refer to the parameters in the state representation (GRU) and emotion classification model. In our experiments we name this model GRU\_Baseline.


\subsection{Training with REINFORCE}
For the action selection module, we used a Monte Carlo Policy Gradient (REINFORCE) \cite{williams1992simple}  action model ($\theta_{a}$) either to $terminate$ or $wait$ for the next frame of the speech utterance. The $terminate$ action triggers the emotion classifier's decision which can be either $neutral$ or $angry$. On the other hand, the $wait$ action does not trigger the decision but waits for the next frame. However, our model triggers the decision after the maximum number of frames as well as the end of the sequence regardless of the selected action.

The action selection module, which is a RL agent, receives two types of rewards, accuracy and latency, which both are terminal rewards. One of the cases, where the agent gets the accuracy reward, is when it chooses the $terminate$ action. Then, the emotion class, which is determined by the emotion classifier is compared with the ground truth label. Thus, the rewards ($r_{acc}$) are \textit{true positive} ($r_{tp}$), \textit{false positive} ($r_{fp}$), \textit{true negative} ($r_{tn}$) and \textit{false negative} ($r_{fn}$) (see Table \ref{table:reward}). When the agent $wait$s more than the maximum number of frames the agent receives a negative reward ($r_{noDec}$) for not triggering the decision (i.e. selecting the $terminate$ action). The agent also receives the latency reward which is
\setlength{\belowdisplayshortskip}{6pt}
\setlength{\abovedisplayshortskip}{6pt}
\begin{equation}
\label{eq.reward_lat}
r_{lat} = \frac{1}{t + 1}
\end{equation}
where $t$ is the termination time frame. In all other cases such as non-terminal steps, the reward is zero (i.e. no intermediate rewards). The total reward function is a summation of accuracy and latency reward:
\setlength{\belowdisplayshortskip}{6pt}
\setlength{\abovedisplayshortskip}{-3pt}
\begin{equation}
\label{eq.total_reward}
r_t = r_{acc} + r_{lat}
\end{equation}
The probability distribution over actions is modeled as a single linear layer with a softmax function:

\begin{equation}
\label{eq.act1}
a_{t} = Softmax(W^a \cdot S_t + b^a)
\end{equation}
where, $W^a$ and $b^a$ ($\theta_a$) are the weight and bias values, and $S_t$ is the averaged hidden state of the GRU. The $Softmax$ function is

\begin{equation}
\label{eq.sotfmax}
Softmax(\alpha_j) = \frac{e^{\alpha_j}}{\sum_{i=1}^{n}e^{\alpha_i}}
\end{equation}

The objective of the RL agent is to maximize the expected return under the agent's policy. 

\setlength{\belowdisplayshortskip}{3pt}
\setlength{\abovedisplayshortskip}{-3pt}

\begin{equation}
\label{eq.returnObjective}
J_{a} (\theta_{a}, \theta_s) = \mathbb{E}_{\pi(a_t|s_t;\ \theta_a, \theta_s)}[R_t]
\end{equation}
where $\pi(a_t|s_t;\ \theta_a, \theta_s)$ is the policy of the agent and $R_t$ is the expected return in each state which is

\begin{equation}
\label{eq.return}
R_t = \sum_{t'=t}^{T} \gamma^{t-t'}r_{t'}
\end{equation}
where $\gamma$ is the discount factor (0.99). To maximize the objective function $J_{a} (\theta_{a}, \theta_s)$, we use the algorithm introduced in \cite{williams1992simple} to approximate the gradient numerically. 

\begin{equation}
\label{eq.returnGradient}
\nabla_{\theta_a, \theta_s} J_{a} (\theta_{a}, \theta_s) \approx \sum_{t=0}^{T} [\nabla_{\theta_a, \theta_s} \ log \pi(a_t|s_t;\theta_a, \theta_s) \ R_t] 
\end{equation}

However, due to the high variance in the gradient signal, we use REINFORCE with baseline reward estimation which needs an extra term $b_t$ to be subtracted from the expected return \cite{williams1992simple}. Therefore, the modified objective function is

\begin{equation}
\label{eq.returnGradientBaseline}
\nabla_{\theta_a, \theta{s}} \ J_{a} (\theta_{a}, \theta_s) \approx \sum_{t=0}^{T} [\nabla_{\theta_a, \theta_s} \ log \pi(a_t|s_t;\theta_a, \theta_s) \ (R_t-b_t)]
\end{equation}
In the next section, we explain the baseline reward estimation.

\subsection{Baseline Reward Estimation}

As discussed in \cite{williams1992simple, P17-1172,zaremba2015reinforcement}, different approaches can be applied to calculate the baseline $b_t$. We applied methods from \cite{gu_learning_2017, mnih2014recurrent}. The baseline is obtained with a linear regression from the hidden state of the GRU:

\begin{equation}
\label{eq.baseline}
b_t = W^b \cdot S_t + b^b
\end{equation}
where $W^b$ and $b^b$ ($\theta_b = [W^b;b^b]$) are the weight and bias values of the baseline model. The objective function to train the baseline parameters is

\begin{equation}
\label{eq.baselineObjective}
J_b (\theta_b)= \mathbb{E}_{\pi(a_t|s_t;\theta_a)} \left [  \sum_{t=0}^{T} (R_t - b_t)^2 \right ]
\end{equation}

It should be noted that we disconnected the gradient signal of the baseline objective function ($\nabla \ J_{\theta_b)}$ to prevent its backpropagation to the hidden state of the GRU. The baseline objective function estimates the expected reward. Therefore, sending its gradient signal to the GRU would eventually change the policy of the model which changes the expected return and thus creates an unstable loop. The expected return with a lower variance is
\begin{equation}
\label{eq.newReturn}
\hat{R_t} = R_t - b_t
\end{equation}
We then applied rescaling introduced by \cite{gu_learning_2017} with a moving average and standard deviation over $\hat{R_t}$, which is

\begin{equation}
\label{eq.newReturnNormalize}
\tilde{R_t} = \frac{\hat{R_t}- \bar{R}}{\sqrt[]{\sigma^2 + \epsilon}}
\end{equation}

Then, the gradient of action selection model (Eq. \ref{eq.returnGradient}) is rewritten as
\begin{equation}
\label{eq.returnGradientNormalizedBaseline}
\nabla_{\theta_a, \theta_s} J_{a} (\theta_{a}, \theta_s) \approx \sum_{t=0}^{T} [\nabla_{\theta_a, \theta_s} \ log \pi(a_t|s_t;\theta_a, \theta_s) \ \tilde{R_t}] \end{equation}


\begin{table}[t]
\vspace{10pt}
\centering
\caption{The accuracy reward values ($r_{acc}$) given to the RL agent}
\label{table:reward}
\resizebox{\columnwidth}{!}{%
\begin{tabular}{cc|c|c|c|l|}
\cline{3-6}
                             &         & \multicolumn{4}{c|}{actions}                                                                                                                                                                              \\ \cline{3-6} 
                             &         & \multicolumn{2}{c|}{terminate}                                                                                          & \multicolumn{2}{c|}{wait}                                                       \\ \cline{3-6} 
                             &         & \begin{tabular}[c]{@{}c@{}}decision:\\ angry\end{tabular} & \begin{tabular}[c]{@{}c@{}}decision:\\ neutral\end{tabular} & \multicolumn{2}{c|}{\begin{tabular}[c]{@{}c@{}}end of\\ utterance\end{tabular}} \\ \hline
\multicolumn{1}{|c|}{Ground} & angry   & $r_{tp} = 1$                                              & $r_{fp} = -1$                                               & \multicolumn{2}{c|}{$r_{noDec} = -1$}                                           \\ \cline{2-6} 
\multicolumn{1}{|c|}{Truth}  & neutral & $r_{fn} = -1$                                             & $r_{tn} = 1$                                                & \multicolumn{2}{c|}{$r_{noDec} = -1$}                                           \\ \hline
\end{tabular}
}
\end{table}

\subsection{Training details}
The total loss function of EmoRL is
\begin{equation}
\label{eq.totalLoss}
J = -J_{a}(\theta_a, \theta_s) \ + \ J_{c}(\theta_c, \theta_s) \ +  \ J_b (\theta_b)
\end{equation}
We used the ADAM optimizer \cite{kingma2014adam} with the learning rate of $10^{-4}$ and a weight decay rate of $10^{-5}$ and used a pre-trained model to improve the learning process. We froze the parameters of the GRU ($\theta_s$) and decision classification ($\theta_c$) for the first 5K episodes after pre-training. The intuition behind the setup was not to jeopardize the pre-trained models due to the large gradient values at the beginning of the training. Our current model was trained with a batch size of 1.

\subsection{Inference}
During training, the action was sampled probabilistically at each time frame ($\pi(a|s_t;\theta_a, \theta_s)$). However, during validation and test, the action with maximum probability was selected: $a = \max_{a'}(\pi(a'|s_t;\theta_a, \theta_s))$. If the $wait$ action was predicted by the model, it moved to the next audio sample. Otherwise, we terminated the processing and compared the prediction of the emotion classification module with the ground truth to estimate performance during the validation phase.

\section{Experimental results}

\subsection{Data}

\par The Interactive Emotional Dyadic Motion Capture dataset IEMOCAP \cite{busso_iemocap:_2008}   contains five recorded sessions of conversations between pairs of actors, one from each gender. The total amount of data is 12 hours of audio-visual information from ten actors annotated with categorical emotion labels (Anger, Happiness, Sadness, Neutral, Surprise, Fear, Frustration and Excitement). Each sample in the corpus is an utterance with an average length of around 6 seconds and is annotated by several annotators. We perform several data filtering steps: we discard samples annotated with three different emotion labels and select only samples that have a consensus of at least two annotators. We select samples annotated only as Anger and Neutral (3,395 utterances overall) as our goal is to evaluate if the person turns into high arousal and negative state. Our goal is to simulate a safety-related scenario, when the robot is able to detect a transition from Neutral to Anger state of a speaker, which can be used for robot's planning and decision making modules.

\subsubsection{Lab setup}
\begin{figure}[t]
\vspace{4pt}
  \includegraphics[width=\linewidth]{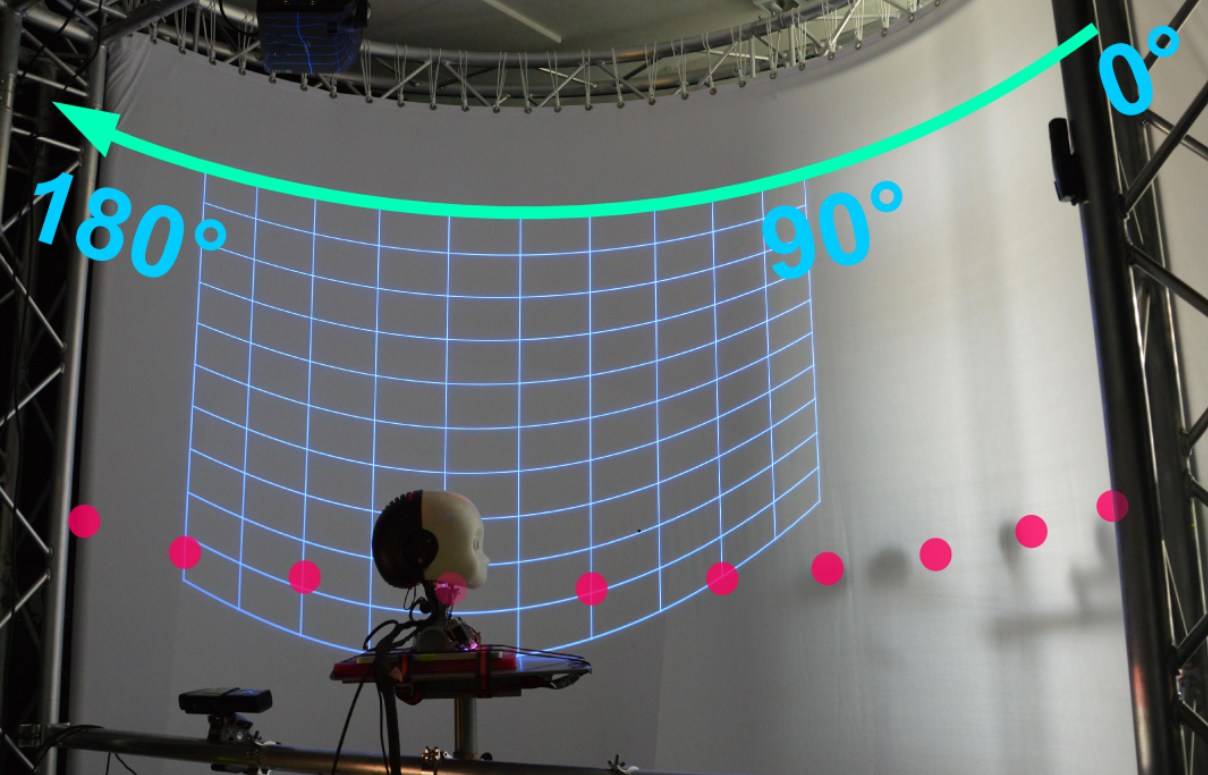}
  \caption{Lab setup of iCub in front of loudspeakers behind a screen \cite{lab_setup_paper}}
  \label{fig:lab_setup}
    \vspace{-10pt}
\end{figure}

Our goal is to simulate a scenario close to a real-life situation. The experimental setup that we use is shown in Fig.\ref{fig:lab_setup}. The setup consists of a humanoid robot head (iCub) immersed in a display to create a virtual reality environment for the robot  \cite{lab_setup_paper}. Speakers are located behind the display between 0\textdegree \ and 180\textdegree \ every 15\textdegree \ along the azimuth plane with the same elevation. The iCub head is 1.6 meters away from the speakers. The setup introduces background noise generated by the projectors, computers, power sources as well as ego noise from the iCub head.

\subsubsection{Dataset recording}
In addition to the clean IEMOCAP dataset we re-recorded this dataset in our lab setup to test the generalization of our method to a human-robot interaction scenario. We play each recording from the IEMOCAP dataset picking a random speaker out of 4 pre-selected speakers in the lab and record a signal from the iCub ear microphones. We use the same annotation for the recorded sample as in the original dataset. As a result we obtain the whole IEMOCAP dataset re-recorded, which has the same acoustic content but is overlaid with several noise types: iCub's ego noise, fan and noise from several PCs present in our lab. We call this dataset IEMOCAP-iCub in our experiments. Therefore, such procedure allows us to test algorithms in a very realistic noise environment on a dataset containing more than 3,000 annotated samples. Recording and annotating such a dataset from scratch in a lab environment would be a time-consuming and error-prone process.

\subsection{Experiments}

\begin{figure}[b]
  \vspace{-10pt}
\includegraphics[width=\linewidth]{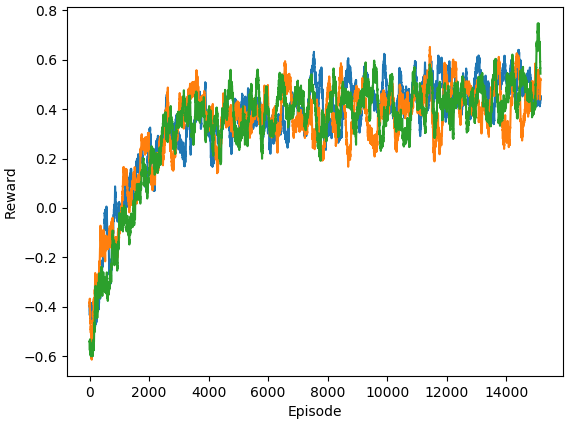}
  \caption{The collected reward by the RL agent during training (three different runs are present corresponding to different cross-validation folds). }
  \label{fig:reward}
  \vspace{0pt}
\end{figure}

\begin{figure}[tbph]
  \includegraphics[width=\linewidth]{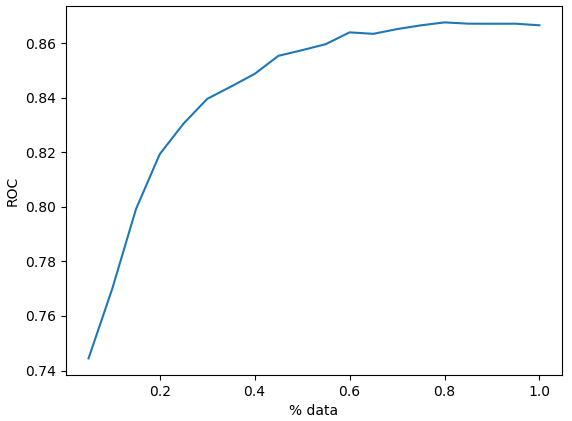}
  \vspace{-20pt}
  \caption{ROC-AUC score (y-axis) of the GRU-based neural model \cite{huang_attention_2016} trained with different ratio of utterance used  (x-axis) from the IEMOCAP-iCub dataset. For example, 0.4 means that we use only first 40\% of the utterance to classify an emotion.}
  \label{fig:baseline_roc_percent}
  \vspace{-20pt}-
\end{figure}
We report accuracy and the area under the receiver operating characteristic curve (AUC-ROC) which is a standard metric for unbalanced binary classifications. In addition to these metrics, we report the achieved speed-up of the models following a leave-one-actor-out routine to assess the generalization ability of the model to the new actors. There are 10 actors present in the dataset and we keep utterances from 9 actors for training and evaluate on the remaining actor. We average results over 10 runs with different random number generator seeds to estimate the variance due to random weight initializations (see Fig.\ref{fig:reward}).

First, we present a graph of dependency between the AUC-ROC metric and ratio of the input sequence used in the analysis (see Fig. \ref{fig:baseline_roc_percent}) and we conclude that performance levels off after 60\% of the sequence is processed. This shows that the emotion classification does not need the whole utterance to achieve best possible performance.

Results for IEMOCAP and IEMOCAP-iCub are present in Table \ref{table:results} and Table \ref{table:results-icub}. EmoRL achieves a 0.86 AUC-ROC score with 1.75x speed-up on average on our recorded IEMOCAP-iCub dataset, while GRU\_Baseline achieves a slightly higher score only with the full-length utterance. EmoRL is as good as GRU\_Baseline if that receives 75\%  of sequence input even though EmoRL reads on average only 57\% of the sequence.
This can be an indicator that our RL agent learns dynamically when it is ready to make predictions.  Moreover, we observe only minor differences in performance of our models on the clean IEMOCAP and IEMOCAP-iCub datasets, which is an indicator that EmoRL can work efficiently even with noise injected.

\begin{table}[tbph]
\centering
\caption{Evaluation Results (IEMOCAP). Rows are sorted w.r.t. speed-up}
\label{table:results}
\resizebox{\columnwidth}{!}{%
\begin{tabular}{|c|c|c|c|c|c|}
\hline
Model       & \begin{tabular}[c]{@{}c@{}} \% of used\\ utterance \end{tabular}            & AUC-ROC                & Accuracy   & \begin{tabular}[c]{@{}c@{}}Relative\\ Latency\end{tabular} & Speed-up       \\ \hline
Most Frequent emotion & - & 0.5                    & 67\%          & 1.0                                                        & -              \\ \hline
GRU\_Baseline \cite{huang_attention_2016} & 10\%   & 0.66$\pm$0.06          &    77.1\%$\pm$8.1\%      & 0.1                                                        & 10x            \\ \hline
GRU\_Baseline \cite{huang_attention_2016}  & 25\%   & 0.67$\pm$0.08   & 78.3\%$\pm$5.8\%  & 0.25  & 4x             \\ \hline
GRU\_Baseline \cite{huang_attention_2016} & 50\%  & 0.71$\pm$0.10  & 82.9\%$\pm$4.9\%           & 0.5                                                        & 2x             \\ \hline
\textbf{EmoRL}    & adaptive    & \textbf{0.89$\pm$0.04} & \textbf{84.9\%$\pm$4.3\%} & \textbf{0.55$\pm$0.2}                                      & \textbf{1.82x} \\ \hline
GRU\_Baseline \cite{huang_attention_2016} & 75\%  & 0.77$\pm$0.11          & 84.8\%$\pm$4.4\%          & 0.75                                                       & 1.3x            \\ \hline
GRU\_Baseline \cite{huang_attention_2016} & 100\%  & 0.90$\pm$0.04          & 85.1\%$\pm$3.9\%         & 1.0                                                        & -              \\ \hline
\end{tabular}
}
\vspace{-10pt}
\end{table}

\begin{table}[tbp]
\vspace{10pt}
\centering
\caption{Evaluation Results (IEMOCAP-iCub). Rows are sorted w.r.t. speed-up}
\vspace{-5pt}
\label{table:results-icub}
\resizebox{\columnwidth}{!}{%
\begin{tabular}{|c|c|c|c|c|c|}
\hline
Model      & \begin{tabular}[c]{@{}c@{}} \% of used\\ utterance \end{tabular}            & AUC-ROC                & Accuracy   & \begin{tabular}[c]{@{}c@{}}Relative\\ Latency\end{tabular} & Speed up       \\ \hline
Most Frequent emotion & - & 0.5                    & 67\%          & 1.0                                                        & -              \\ \hline
GRU\_Baseline \cite{huang_attention_2016} & 10\%  & 0.74$\pm$0.05          & 77.0\%$\pm$7.1\%           & 0.1                                                        & 10x            \\ \hline
GRU\_Baseline \cite{huang_attention_2016}  & 25\%  & 0.75$\pm$0.06          & 79.5\%$\pm$6.4          & 0.25                                                       & 4x             \\ \hline
GRU\_Baseline \cite{huang_attention_2016} & 50\%  & 0.77$\pm$0.08          & 81.4\%$\pm$5.7\%        & 0.5                                                        & 2x             \\ \hline
\textbf{EmoRL}  & adaptive      & \textbf{0.86$\pm$0.05} & \textbf{82.5\%$\pm$5.1\%} & \textbf{0.57$\pm$0.24}                                     & \textbf{1.75x} \\ \hline
GRU\_Baseline \cite{huang_attention_2016} & 75\%  & 0.81$\pm$0.07          & 82.0\%$\pm$6.1\%          & 0.75                                                       & 1.3x           \\ \hline
GRU\_Baseline \cite{huang_attention_2016} & 100\% & 0.87$\pm$0.04          & 82.8\%$\pm$5.5\%         & 1.0                                                        & -              \\ \hline
\end{tabular}
}
\vspace{-10pt}
\end{table}

\section{CONCLUSIONS}
We presented a model for acoustic emotion recognition, which is able to decide adaptively to emit the prediction as early as possible keeping a high level of accuracy. To the best of our knowledge, our model is the first implementation using the policy gradient for early emotion classification. Our model  is able to distinguish angry emotion from neutral on average 1.75 times earlier and achieves similar performance compared to the GRU\_Baseline (which uses the whole utterance). We especially selected the angry emotion due to its potential applications in safety scenarios and its urgency among other emotions for an early detection. 

Our model keeps the performance levels comparable in the clean IEMOCAP and noisy IEMOCAP-iCub datasets. A reason for this could be that in our architecture, the action selection model does not directly obtain the emotion classification output and only learns to optimize with the terminal reward. 

Our model can also be applied to other sequence classification tasks such as gesture recognition. As future work we want to include other modalities, like vision, to improve the performance of our model. Moreover, including other emotions such as \textit{happiness} and \textit{sadness} can extend our model applications for instance to conduct a dialogue.

\addtolength{\textheight}{-12cm}   




\section*{ACKNOWLEDGMENT}

This project has received funding from the European Union's Horizon 2020 research and innovation programme under the Marie Sklodowska-Curie grant agreement No 642667 (SECURE) and Crossmodal Learning (TRR 169). The authors would like to thank Erik Strahl for his support with the experimental setup and Julia Lakomkina for her help with illustrations.


\bibliography{maz_bib.bib,Zotero_egor}

\begin{thebibliography}{10}
\providecommand{\url}[1]{#1}
\csname url@samestyle\endcsname
\providecommand{\newblock}{\relax}
\providecommand{\bibinfo}[2]{#2}
\providecommand{\BIBentrySTDinterwordspacing}{\spaceskip=0pt\relax}
\providecommand{\BIBentryALTinterwordstretchfactor}{4}
\providecommand{\BIBentryALTinterwordspacing}{\spaceskip=\fontdimen2\font plus
\BIBentryALTinterwordstretchfactor\fontdimen3\font minus
  \fontdimen4\font\relax}
\providecommand{\BIBforeignlanguage}[2]{{%
\expandafter\ifx\csname l@#1\endcsname\relax
\typeout{** WARNING: IEEEtran.bst: No hyphenation pattern has been}%
\typeout{** loaded for the language `#1'. Using the pattern for}%
\typeout{** the default language instead.}%
\else
\language=\csname l@#1\endcsname
\fi
#2}}
\providecommand{\BIBdecl}{\relax}
\BIBdecl

\bibitem{jozefowicz_exploring_2016}
R.~Jozefowicz, O.~Vinyals, M.~Schuster, N.~Shazeer, and Y.~Wu, ``Exploring the
  {Limits} of {Language} {Modeling},'' \emph{arXiv:1602.02410 [cs]}, Feb. 2016,
  arXiv: 1602.02410.

\bibitem{radford_learning_2017}
\BIBentryALTinterwordspacing
A.~Radford, R.~Jozefowicz, and I.~Sutskever, ``Learning to {Generate} {Reviews}
  and {Discovering} {Sentiment},'' \emph{arXiv:1704.01444 [cs]}, Apr. 2017,
  arXiv: 1704.01444. [Online]. Available: \url{http://arxiv.org/abs/1704.01444}
\BIBentrySTDinterwordspacing

\bibitem{hannun_deep_2014}
A.~Hannun, C.~Case, J.~Casper, B.~Catanzaro, and et~al., ``Deep {Speech}:
  {Scaling} up end-to-end speech recognition,'' \emph{CoRR}, vol.
  abs/1412.5567, Dec. 2014.

\bibitem{wu_googles_2016}
Y.~Wu, M.~Schuster, Z.~Chen, Q.~V. Le, M.~Norouzi, W.~Macherey, M.~Krikun, and
  et~al, ``Google's {Neural} {Machine} {Translation} {System}: {Bridging} the
  {Gap} between {Human} and {Machine} {Translation},'' \emph{arXiv:1609.08144
  [cs]}, Sep. 2016, arXiv: 1609.08144.

\bibitem{fayek_evaluating_2017}
H.~M. Fayek, M.~Lech, and L.~Cavedon, ``Evaluating deep learning architectures
  for {Speech} {Emotion} {Recognition},'' \emph{Neural Networks}, vol. vol. 92,
  pp. 60--68, Jan. 2017.

\bibitem{huang_attention_2016}
C.-W. Huang and S.~S. Narayanan, ``Attention {Assisted} {Discovery} of
  {Sub}-{Utterance} {Structure} in {Speech} {Emotion} {Recognition},'' in
  \emph{Proceedings of {Interspeech}}, Sep. 2016, pp. 1387--1391.

\bibitem{lee_high-level_2015}
J.~Lee and I.~Tashev, ``High-level feature representation using recurrent
  neural network for speech emotion recognition.'' in \emph{{INTERSPEECH}},
  2015, pp. 1537--1540.

\bibitem{longpre2016way}
S.~Longpre, S.~Pradhan, C.~Xiong, and R.~Socher, ``A way out of the odyssey:
  Analyzing and combining recent insights for lstms,'' \emph{arXiv preprint
  arXiv:1611.05104}, 2016.

\bibitem{krizhevsky_imagenet_2012}
\BIBentryALTinterwordspacing
A.~Krizhevsky, I.~Sutskever, and G.~E. Hinton, ``{ImageNet} {Classification}
  with {Deep} {Convolutional} {Neural} {Networks},'' in \emph{Advances in
  {Neural} {Information} {Processing} {Systems} 25}, F.~Pereira, C.~J.~C.
  Burges, L.~Bottou, and K.~Q. Weinberger, Eds.\hskip 1em plus 0.5em minus
  0.4em\relax Curran Associates, Inc., 2012, pp. 1097--1105. [Online].
  Available:
  \url{http://papers.nips.cc/paper/4824-imagenet-classification-with-deep-convolutional-neural-networks.pdf}
\BIBentrySTDinterwordspacing

\bibitem{simonyan_very_2014}
K.~Simonyan and A.~Zisserman, ``Very deep convolutional networks for
  large-scale image recognition,'' \emph{CoRR}, vol. abs/1409.1556, 2014.

\bibitem{ghosh_representation_2016}
S.~Ghosh, E.~Laksana, L.-P. Morency, and S.~Scherer, ``Representation
  {Learning} for {Speech} {Emotion} {Recognition}.'' \emph{INTERSPEECH}, pp.
  3603--3607, 2016.

\bibitem{trigeorgis_adieu_2016}
G.~Trigeorgis, F.~Ringeval, R.~Brueckner, E.~Marchi, M.~A. Nicolaou,
  B.~Schuller, and S.~Zafeiriou, ``Adieu features? {End}-to-end speech emotion
  recognition using a deep convolutional recurrent network,'' in
  \emph{Acoustics, {Speech} and {Signal} {Processing} ({ICASSP}), 2016 {IEEE}
  {International} {Conference} on}.\hskip 1em plus 0.5em minus 0.4em\relax
  IEEE, 2016, pp. 5200--5204.

\bibitem{P17-1172}
A.~W. Yu, H.~Lee, and Q.~Le, ``Learning to skim text,'' in \emph{Proceedings of
  the 55th Annual Meeting of the Association for Computational Linguistics
  (Volume 1: Long Papers)}.\hskip 1em plus 0.5em minus 0.4em\relax Association
  for Computational Linguistics, 2017, pp. 1880--1890.

\bibitem{hochreiter_long_1997}
S.~Hochreiter and J.~Schmidhuber, ``Long {Short}-{Term} {Memory},''
  \emph{Neural Comput.}, vol.~9, no.~8, pp. 1735--1780, Nov. 1997.

\bibitem{williams1992simple}
R.~J. Williams, ``Simple statistical gradient-following algorithms for
  connectionist reinforcement learning,'' \emph{Machine learning}, vol.~8, no.
  3-4, pp. 229--256, 1992.

\bibitem{shen_reasonet:_2017}
\BIBentryALTinterwordspacing
Y.~Shen, P.-S. Huang, J.~Gao, and W.~Chen, ``{ReasoNet}: {Learning} to {Stop}
  {Reading} in {Machine} {Comprehension},'' in \emph{Proceedings of the 23rd
  {ACM} {SIGKDD} {International} {Conference} on {Knowledge} {Discovery} and
  {Data} {Mining}}, ser. {KDD} '17.\hskip 1em plus 0.5em minus 0.4em\relax New
  York, NY, USA: ACM, 2017, pp. 1047--1055. [Online]. Available:
  \url{http://doi.acm.org/10.1145/3097983.3098177}
\BIBentrySTDinterwordspacing

\bibitem{ranzato_sequence_2016}
M.~Ranzato, S.~Chopra, M.~Auli, and W.~Zaremba, ``Sequence {Level} {Training}
  with {Recurrent} {Neural} {Networks},'' \emph{International Conference on
  Learning Representations}, 2016.

\bibitem{gu_learning_2017}
J.~Gu, G.~Neubig, K.~Cho, and V.~O.~K. Li, ``\BIBforeignlanguage{English
  (US)}{Learning to translate in real-time with neural machine translation},''
  in \emph{\BIBforeignlanguage{English (US)}{15th {Conference} of the
  {European} {Chapter} of the {Association} for {Computational}
  {Linguistics}}}.\hskip 1em plus 0.5em minus 0.4em\relax Association for
  Computational Linguistics (ACL), 2017.

\bibitem{eyben_recent_2013}
F.~Eyben, F.~Weninger, F.~Gross, and B.~Schuller, ``Recent {Developments} in
  {openSMILE}, the {Munich} {Open}-source {Multimedia} {Feature} {Extractor},''
  in \emph{Proceedings of the 21st {ACM} {International} {Conference} on
  {Multimedia}}, ser. {MM} '13.\hskip 1em plus 0.5em minus 0.4em\relax New
  York, NY, USA: ACM, 2013, pp. 835--838.

\bibitem{bahdanau_neural_2015}
D.~Bahdanau, K.~Cho, and Y.~Bengio, ``Neural {Machine} {Translation} by
  {Jointly} {Learning} to {Align} and {Translate},'' \emph{ICLR}, 2015.

\bibitem{zaremba2015reinforcement}
W.~Zaremba and I.~Sutskever, ``Reinforcement learning neural turing
  machines-revised,'' \emph{arXiv preprint arXiv:1505.00521}, 2015.

\bibitem{mnih2014recurrent}
V.~Mnih, N.~Heess, A.~Graves \emph{et~al.}, ``Recurrent models of visual
  attention,'' in \emph{Advances in neural information processing systems},
  2014, pp. 2204--2212.

\bibitem{kingma2014adam}
D.~Kingma and J.~Ba, ``Adam: A method for stochastic optimization,'' in
  \emph{3rd International Conference for Learning Representations}, San Diego,
  2015.

\bibitem{busso_iemocap:_2008}
C.~Busso, M.~Bulut, C.-C. Lee, A.~Kazemzadeh, E.~Mower, S.~Kim, J.~N. Chang,
  S.~Lee, and S.~S. Narayanan, ``\BIBforeignlanguage{en}{{IEMOCAP}: interactive
  emotional dyadic motion capture database},''
  \emph{\BIBforeignlanguage{en}{Language Resources and Evaluation}}, vol.~42,
  no.~4, p. 335, Dec. 2008.

\bibitem{lab_setup_paper}
J.~Bauer, J.~D\'avila-Chac\'on, E.~Strahl, and S.~Wermter, ``Smoke and mirrors
  - virtual realities for sensor fusion experiments in biomimetic robotics,''
  in \emph{Proceedings of the 2012 IEEE International Conference on Multisensor
  Fusion and Information Integration (MFI 2012)}, Hamburg, DE, Sep 2012, pp.
  114--119.

\end{thebibliography}

\end{document}